\documentclass{article}

\usepackage[preprint]{cs224r_2025}
\usepackage[utf8]{inputenc} 
\usepackage[T1]{fontenc}    
\usepackage{hyperref}       
\usepackage{url}            
\usepackage{booktabs}       
\usepackage{amsfonts}       
\usepackage{nicefrac}       
\usepackage{microtype}      
\usepackage{xcolor}         
\usepackage{amsmath}
\usepackage{graphicx}       
\usepackage{lipsum}         
\usepackage{float}
\usepackage{makecell}
\usepackage{algorithm}
\usepackage{algpseudocode}
\usepackage{amsmath}
\usepackage{url} 
\usepackage{adjustbox}   

\title{Bellman-Consistent Trajectory RL for LLM Arithmetic Reasoning}

\author{%
  Lianfa Li$^{1,2}$ \\
  $^{1}$Department of Computer Science, Stanford University \\
  $^{2}$LREIS, Institute of Geographical Sciences and Natural Resources Research, \\
  \hspace{0.6em}Chinese Academy of Sciences \\
  \texttt{lianfali@stanford.edu}, \texttt{lilf@lreis.ac.cn}
}

\begin{document}

\begin{titlepage}
\begin{center}
\Large\textbf{Extended Abstract}
\end{center}
\vspace{10pt}

\paragraph{Motivation} Training Large Language Models (LLMs) for multi-step arithmetic reasoning remains a significant challenge due to sparse terminal rewards and the lack of intermediate guidance. In tasks like Countdown, where models must generate sequential operations to reach a target number, rewards are only allocated at the sequence's end. Standard reinforcement learning approaches assign credit solely to the final step, leaving intermediate reasoning steps without meaningful learning signals. This project addresses this critical bottleneck: the fundamental mismatch between sparse terminal rewards and the dense, step-wise supervision required to learn effective sequential strategies and intermediate representations.

\paragraph{Method} To overcome this, we propose a Bellman-consistent trajectory RL framework with three core objectives: (1) transforming sparse terminal rewards into dense token-level supervision via Trajectory Bellman Residual Minimization (TBRM); (2) reducing gradient variance during optimization using a Reinforce Leave-One-Out (RLOO) or Group Relative Policy Optimization (GRPO) strategy; and (3) enabling effective intermediate step learning via hindsight goal relabeling. We evaluate our framework on the Countdown task, conducting ablation studies to quantify each component's contribution to reasoning accuracy and training stability.

\paragraph{Implementation} Our method builds on TBRM, utilizing soft reward and Q-functions to provide analytical solutions that allow a single LLM to serve as the Q, V, and policy functions, significantly reducing optimization complexity. The soft performance difference lemma guarantees that TBRM satisfies the optimal BRM solution, preserving token-level consistent optimization for the final reward. Additionally, we prove that the RLOO grouping strategy considerably reduces the variance of soft Bellman residuals, enhancing training stability. Finally, hindsight goal relabeling is applied to failure samples to strengthen intermediate reasoning steps. We evaluate our approach on the Countdown task, benchmarking against standard policy gradient methods and ablated baselines.

\paragraph{Results} The TBRM-RLOO framework significantly reduces gradient variance compared to vanilla TBRM or TBRM-GRPO, leading to superior training stability and performance, with hindsight goal relabeling further accelerating learning efficiency. TBRM-RLOO improves rollout accuracy from 0.61 to 0.73 (+20\%), Pass@1 from 0.55 to 0.56, and Pass@16 from 0.74 to 0.84 (+14\%). Incorporating goal relabeling further boosts Pass@1 to 0.64. RLOO grouping consistently outperforms the GRPO strategy. Compared to SFT, IPO/DPO, RLOO, and vanilla TBRM, our method achieves the top overall Pass@16 performance at 84.0\%, outperforming the RLOO baseline by 10.0\% (absolute) and SFT by 8.0\%. It also exhibits superior single-sample efficiency, dominating the low-compute regime (56.1-64.0\% Pass@1 within 100 steps), and successfully circumvents the 72.0\% plateau encountered by existing methods via test-time scaling.

\paragraph{Discussion} Our results demonstrate that a Bellman-consistent trajectory RL method combined with RLOO variance reduction and hindsight goal relabeling significantly improves LLM performance on multi-step arithmetic tasks. The TBRM framework effectively densifies sparse rewards into token-level supervision, while the RLOO strategy stabilizes training by mitigating optimization variance stemming from prompt or problem difficulty. Hindsight goal relabeling enriches data and accelerates intermediate step learning. While constrained by model complexity and broader task generalization, our findings suggest that this framework provides a robust blueprint for training LLMs on complex, sparsely rewarded reasoning tasks.

\paragraph{Conclusion} We introduced a Bellman-consistent trajectory RL method that successfully transforms sparse rewards into dense token-level supervision via TBRM, stabilizes optimization via RLOO variance reduction, and enhances sample efficiency through hindsight goal relabeling. Our framework achieves top-tier results on the Countdown task, balancing exploration and exploitation while successfully bypassing traditional performance plateaus. Future work will investigate the scalability of this approach across diverse multi-step reasoning domains.

\end{titlepage}

\maketitle

\begin{abstract}
To address sparse terminal rewards in LLM arithmetic reasoning, we propose a Bellman-consistent trajectory RL method. Our framework introduces Trajectory Bellman Residual Minimization (TBRM) to transform sparse rewards into dense token-level signals. We integrate a Reinforce Leave-One-Out (RLOO) group strategy to reduce gradient variance of Bellman residuals and use hindsight goal relabeling to accelerate intermediate learning. Evaluated on the Countdown task, our method alleviates optimization bottlenecks and stabilizes gradient variance significantly better than GRPO. The TBRM-RLOO framework boosts Pass@1 to 0.55--0.64 and achieves a state-of-the-art Pass@16 accuracy of 76.0--84.0\%, outperforming the SFT, IPO/DPO, and RLOO baselines. Our approach demonstrates exceptional single-sample efficiency and successfully circumvents standard performance plateaus via test-time scaling.
\end{abstract}

\section{Introduction}
Large Language Models (LLMs) have demonstrated remarkable capabilities across a broad spectrum of natural language processing tasks. However, multi-step arithmetic reasoning remains a significant frontier and a core challenge \citep{li2025system}. Tasks such as the Countdown game require the model to plan, execute, and verify a precise sequence of arithmetic operations to arrive at a target number \citep{herr2025llm,katz2025seemingly,lian2025reinforcement}. In these complex, multi-step settings, Reinforcement Learning (RL) has emerged as a core paradigm for enhancing reasoning capabilities via reward-driven optimization \citep{ouyang2022training, shao2024deepseekmath, guo2025deepseek}. Existing approaches typically fall into two categories: online policy-gradient methods (e.g., REINFORCE, PPO, RLOO) that optimize token-level policies directly from trajectory rollouts, and preference-based optimization methods (e.g., DPO, IPO) that align models offline using pairwise critique datasets.

Despite these algorithmic advances, training LLMs for sequential arithmetic reasoning is bottlenecked by three major challenges:
\begin{enumerate}
\item \textbf{Sparse Terminal Rewards vs. Dense Supervision:} Standard RL approaches assign credit only at the final step based on the correctness of the final output. This leaves intermediate reasoning tokens without local, step-wise learning signals, making it difficult for the model to learn effective intermediate representations \citep{lightman2024let}.
\item \textbf{High Gradient Variance:} Large action spaces and the structural inconsistency between policy and parameter spaces—both inherent to token-level generation—lead to severe training instability and high gradient variance during online trajectory rollouts \citep{schulman2017proximal}. 
\item \textbf{Poor Sample Efficiency:} Standard online RL requires a prohibitive volume of environmental interactions, heavily taxing computational resources when high-quality trajectories are rare \citep{shao2024deepseekmath}.
\end{enumerate}

To address these fundamental limitations, this project introduces a \textbf{Bellman-consistent trajectory RL method} tailored for LLM arithmetic reasoning. Our framework leverages \textbf{Trajectory Bellman Residual Minimization (TBRM)} \citep{yuan2026trajectory} to transform sparse terminal rewards into dense, token-level supervision signals. By shifting the optimization objective from individual transitions to the entire rollout, TBRM provides a value-based method for reasoning that operates effectively with just a single rollout, bypassing the need for separate critic networks or complex importance ratios.

To stabilize training dynamics and maximize data utility, we integrate a grouping statistical strategy drawing on \textbf{Reinforce Leave-One-Out (RLOO)} \citep{ahmadian2024back} or Group Relative Policy Optimization (GRPO) \citep{schulman2017proximal}, which drastically reduces variance during gradient optimization via multi-sample group baselines. Finally, we employ hindsight goal relabeling \citep{andrychowicz2017hindsight} to replay high-reward historical trajectories, accelerating the learning of intermediate reasoning steps from sparse feedback. 

We evaluate our proposed Bellman-consistent trajectory RL framework on the rigorous Countdown task. Empirical results demonstrate that our method significantly alleviates optimization bottlenecks and stabilizes gradient variance compared to SFT, IPO/DPO, and RLOO baselines. Specifically, our approach boosts Pass@1 to 55.0--64.0\% and achieves a state-of-the-art Pass@16 accuracy of 76.0--84.0\%. This demonstrates exceptional single-sample efficiency—driven by TBRM and hindsight goal relabeling—while successfully circumventing standard performance plateaus via test-time scaling. Ultimately, this framework provides a good blueprint for training LLMs on complex, sparsely rewarded reasoning tasks, with potential applications extending well beyond arithmetic to broader multi-step reasoning domains.

\section{Related Work}
\subsection{RLHF and Preference Alignment for Mathematical Reasoning}
Reinforcement Learning from Human (or AI) Feedback has become the standard for aligning LLMs with complex downstream tasks \citep{rlhf}. Early frameworks relied primarily on token-level policy gradient methods like REINFORCE \citep{williams1992simple} and Proximal Policy Optimization (PPO) \citep{schulman2017proximal}. While PPO optimizes token policies directly from online trajectory rollouts, it fundamentally suffers from a lack of intermediate feedback and high gradient variance during long rollouts, especially when reward signals are sparse.

To circumvent the instability and computational overhead of training separate online reward and critic models, preference-based optimization methods such as Direct Preference Optimization (DPO) \citep{rafailov2023direct} and Identity Policy Optimization (IPO)\citep{ahmadian2024back} have been introduced. These methods bypass online sampling entirely by directly optimizing a preference loss over an offline dataset of pairwise comparisons. However, because offline preference methods lack exploratory online rollouts, they can struggle to discover novel, valid mathematical paths in challenging reasoning tasks.

More recently, group-based multi-sample baseline methods such as 
Reinforce Leave-One-Out (RLOO)~\citep{ahmadian2024back} and Group 
Relative Policy Optimization (GRPO)~\citep{shao2024deepseekmath} 
have gained traction. These algorithms reduce policy gradient variance by computing advantages relative to a group of outputs sampled from the same prompt, stabilizing training without requiring a separate critic network. However, they still inherently rely on sparse terminal rewards and lack native token-level credit assignment. In a different direction, \citet{zhu2025flowrl} propose FlowRL, which replaces reward maximization with flow-based distribution matching, learning to balance the full reward distribution across trajectories rather than optimizing a scalar signal directly.

\subsection{Bellman Residual Minimization}
Value-based reinforcement learning relies heavily on the Bellman equation to propagate rewards backward through time \citep{baird1995residual, sutton1998reinforcement}. Traditional Bellman Residual Minimization (BRM) aims to minimize the squared error of the Bellman equation across individual transitions \citep{antos2008learning}. However, classical BRM is severely limited by the fundamental double sampling problem under non-linear transitions \citep{dann2014policy}. Specifically, obtaining an unbiased gradient of the Bellman residual requires two independent samples of the next state from the same current state-action pair. In environments with non-linear transitions or vast token action spaces, single-sample estimates become heavily biased, which severely limits the stability of stochastic residual optimization \citep{baird1995residual,dann2014policy}.

Our work departs from traditional BRM by utilizing \textbf{Trajectory Bellman Residual Minimization (TBRM)} that has been proposed in \citet{yuan2026trajectory}. Instead of minimizing errors over isolated transitions, TBRM shifts the objective to the scale of the entire rollout trajectory. This formulation allows us to derive a simple, value-based method explicitly tuned for LLM reasoning text. By evaluating consistency across the complete trajectory, TBRM effectively sidesteps the double sampling problem, allowing for stable optimization using only a single rollout sample per prompt without a dedicated critic network.

\subsection{Hindsight Goal Relabeling}

Originally introduced in goal-conditioned reinforcement learning, Hindsight Goal Relabeling (HGR) addresses the challenge of sparse rewards by treating failed trajectories as successful rollouts for alternative goals \citep{andrychowicz2017hindsight}. By dynamically relabeling unsuccessful attempts with the actually achieved states, HGR transforms counterfactual experiences into valuable training data, thereby maximizing sample efficiency.

In this work, we adapt HGR to the domain of multi-step arithmetic reasoning in LLMs. Specifically, when a model generates an incorrect final answer, we relabel the intermediate multi-step derivation with its actual mathematical outcome, effectively converting a failed rollout into a successful demonstration for a modified problem context. This mechanism enables the model to extract meaningful gradient signals from its own errors, accelerating the optimization of intermediate reasoning steps when the ground-truth reward is highly sparse. While HGR is heavily utilized in continuous control and robotic manipulation, its application to structured language generation and complex reasoning tasks is a novel extension that yields significant gains in convergence speed and final task accuracy.

\section{Method}
\begin{figure}[htbp]
  \centering
  \includegraphics[width=1\linewidth]{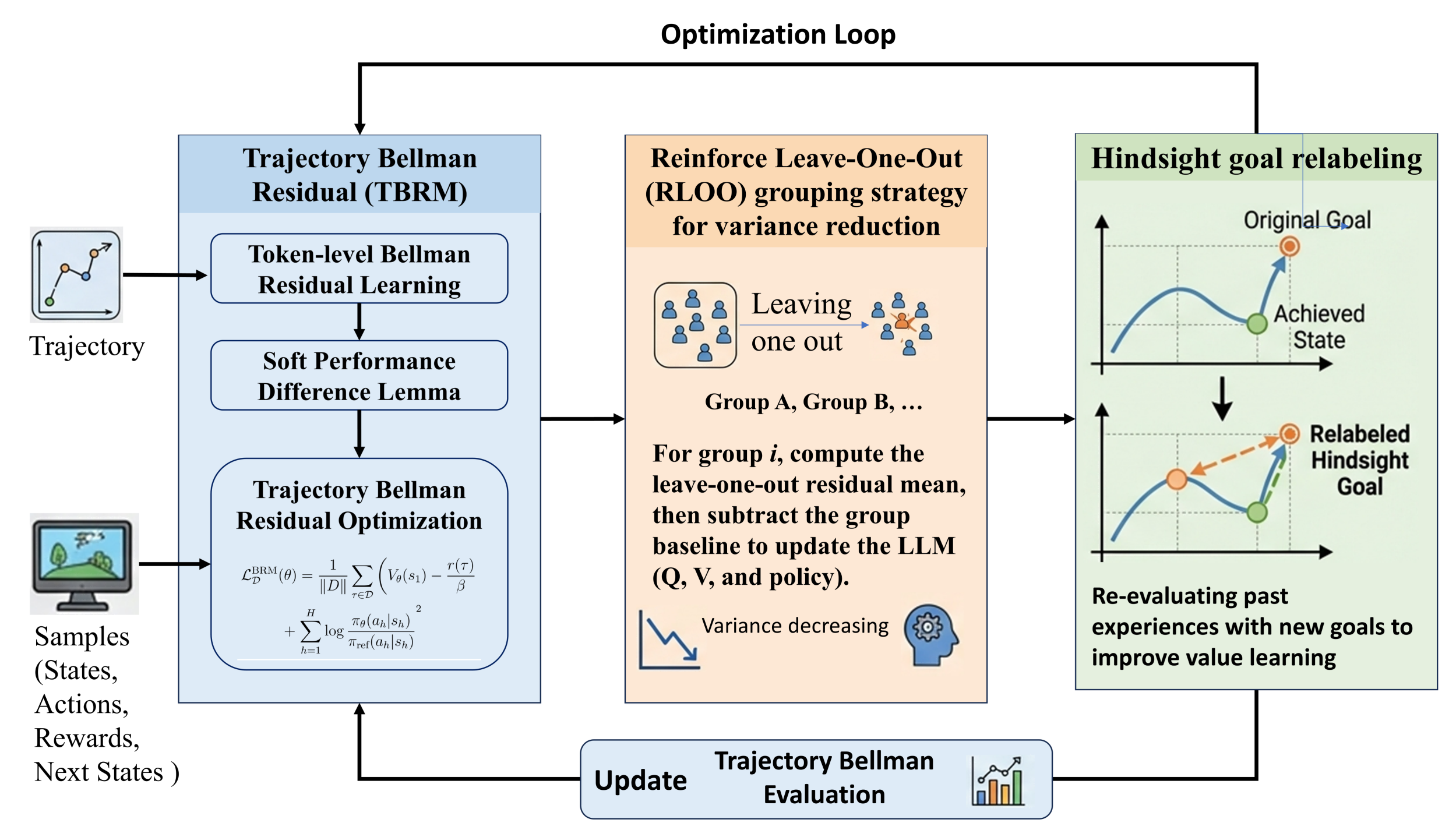}
  \caption{Overview of the Bellman-consistent trajectory RL framework, detailing the coupling of Trajectory Bellman Residual Minimization (TBRM) with group-wise RLOO baseline adjustment and hindsight goal relabeling.}
  \label{fig:sysframe}
\end{figure}

The proposed method introduces a unified value-based framework for reinforcing multi-step language model reasoning without relying on separate critic architectures or complex importance sampling ratios. As illustrated in Figure \ref{fig:sysframe}, the framework consists of three tightly integrated components: 
\begin{enumerate}
   \renewcommand{\labelenumi}{\arabic{enumi})}
    \item \textbf{Trajectory Bellman Residual Minimization (TBRM):} A value-based objective that maps token logits directly to action-values, converting sparse terminal signals into dense token-level supervision while preserving token-level consistency.
    \item \textbf{Reinforce Leave-One-Out (RLOO) Grouping Strategy:} A variance-reduction technique tailored to group-wise Bellman residuals to stabilize policy updates.
    \item \textbf{Hindsight Goal Relabeling (HGR):} An experience augmentation strategy that transforms failed rollouts into high-value successful trajectories to accelerate intermediate learning.
\end{enumerate}

\subsection{Trajectory Bellman Residual Minimization (TBRM)} 
Under the Kullback-Leibler (KL) regularized reinforcement learning paradigm with a reference policy $\pi_{\text{ref}}$, the TBRM objective maximizes the regularized trajectory-level return \citep{yuan2026trajectory}:
\begin{align}
J_\beta(\pi) & = \beta \mathbb{E}_{\tau \sim \pi} \left[R_\beta(\tau)-\log \pi(\tau) \right] \nonumber\\
&= \beta \mathbb{E}_{\tau \sim \pi} \left[\sum_{h=1}^H \Big( R_\beta(s_h,a_h)-\log \pi(a_h|s_h) \Big) \right]
\end{align}
where $\beta > 0$ is the temperature hyperparameter, and $R_\beta(s_h,a_h)$ is the token-level soft reward function:
\begin{equation}
R_\beta(s_h,a_h)=\frac{r(s_h,a_h)}{\beta}+\log \pi_{\text{ref}}(a_h|s_h)
\end{equation}

The optimal policy maximizing the regularized objective, $\pi_\beta^* = \arg \max_\pi J_\beta(\pi)$, can be derived analytically in closed form as a reweighted softmax distribution. Depending on whether the reference policy is explicitly decoupled or implicitly absorbed, the solution can be equivalently expressed in two standard forms \citep{yuan2026trajectory}:
\begin{align*}
  \pi_\beta^*(a_h|s_h) \propto \pi_{\text{ref}}(a_h|s_h) \exp \left( Q_r^*(s_h,a_h)/\beta \right)
  \text{  or  } \pi_\beta^*(a_h|s_h) \propto \exp \left(Q_{R_\beta}^*(s_h,a_h) \right)
\end{align*}
where $Q_{R_\beta}^*(s_h,a_h) = Q_r^*(s_h,a_h) + \beta \log \pi_{\text{ref}}(a_h|s_h)$ represents the optimal action-value function.

Consequently, the soft Bellman operator $\mathcal{T}_\beta$ for the $Q$ function can be expressed in closed form as:
\begin{align}
(\mathcal{T}_\beta Q)(s,a) : &= R_\beta(s,a) + \gamma \mathbb{E}_{s' \sim \mathcal{P}(\cdot|s,a)}\big[ V_Q(s') \big] \nonumber \\
&=  R_\beta(s,a) + \gamma \log \sum_{a' \in \mathcal{A}} \exp \left( Q(s',a') \right) 
\end{align}
where $s'$ is the deterministic next state resulting from taking action $a$ in state $s$, and $\gamma$ is the discount factor (often set to $1$ in finite-horizon token generation).

Under the autoregressive language modeling setting, state transitions are deterministic: the state $s_h$ consists of the prompt concatenated with all generated tokens up to step $h-1$, and the action $a_h$ is the next token in the vocabulary $\mathcal{A}$. We parameterize the action-value function $Q_\theta$, the value function $V_\theta$, and the resulting policy $\pi_\theta$ directly via the model's unnormalized log-probabilities (logits) without an independent critic network:
\begin{align*}
  Q_\theta(s,a):= \text{logit}_{\theta}(s,a); 
  V_\theta(s):= \log \sum_{a' \in \mathcal{A}} \exp(\text{logit}_\theta(s,a')); 
  \log\pi_\theta(a|s)= Q_\theta(s,a) - V_\theta(s)
\end{align*}
where $V_\theta(s)$ is the log-sum-exp operator over the token vocabulary. 

In deterministic environments, classical Bellman Residual Minimization (BRM) minimizes the squared per-step temporal difference error over a trajectory dataset $\mathcal{D}$:
\begin{align}
&\mathcal{L}_\mathcal{D}^{\text{BRM}}(\theta) 
= \frac{1}{|\mathcal{D}|}\sum_{\tau \in \mathcal{D}}\sum_{h=1}^H 
   \left(\text{logit}_\theta(s_h,a_h)
   - (\mathcal{T}_\beta \text{logit}_\theta)(s_h,a_h) \right)^2 \nonumber \\
&= \frac{1}{|\mathcal{D}|}\sum_{\tau \in \mathcal{D}}\sum_{h=1}^H 
   \left(\text{logit}_\theta(s_h,a_h)
   - \frac{r(s_h,a_h)}{\beta} 
   - \log \pi_{\text{ref}}(a_h|s_h) 
   - \log \sum_{a' \in \mathcal{A}} \exp(\text{logit}_\theta(s_{h+1},a'))
   \right)^2
\end{align}
However, directly optimizing $\mathcal{L}_\mathcal{D}^{\text{BRM}}(\theta)$ fails in language model post-training because the process reward $r(s_h, a_h)$ is typically unavailable or extremely sparse, only appearing as a terminal signal $r(\tau)$ at the final token $h=H$. 

To bypass this limitation, \citet{yuan2026trajectory} lift the step-wise residual to a trajectory-level objective:
\begin{align}
\mathcal{L}_\mathcal{D}^{\text{TBRM}}(\theta)
&= \frac{1}{|\mathcal{D}|}\sum_{\tau \in \mathcal{D}}
   \left( \sum_{h=1}^H \left[ \text{logit}_\theta(s_h,a_h) - (\mathcal{T}_\beta \text{logit}_\theta)(s_h,a_h) \right] \right)^2 \nonumber \\
&= \frac{1}{|\mathcal{D}|}\sum_{\tau \in \mathcal{D}}
   \left( V_\theta(s_1)
   - \frac{r(\tau)}{\beta}
   + \sum_{h=1}^H \log \frac{\pi_\theta(a_h|s_h)}{\pi_{\text{ref}}(a_h|s_h)}
   \right)^2
\end{align}
TBRM eliminates the need for intermediate token-level reward labeling or value scoring, using only the final trajectory outcome $r(\tau)$. Crucially, the formulation satisfies the solution of the original token-level BRM framework, maintaining token-level consistent optimization for the final reward while propagating the sparse terminal reward back into dense token-level supervision signals.

The theoretical validity of optimizing the policy via trajectory-level residuals is justified by the soft performance difference lemma. Let $\hat{\pi}(a|s) = \frac{\exp(Q(s,a))}{\sum_{a' \in \mathcal{A}}\exp(Q(s,a'))}$ be the soft-greedy policy induced by the learned value function. The suboptimality gap between the globally optimal policy $\pi^*$ and $\hat{\pi}$ is strictly upper-bounded by the expected trajectory Bellman residual:
\begin{equation}
J_\beta(\pi^*) - J_\beta(\hat{\pi}) 
\leq 2\beta \max_{\pi \in \{\pi^*, \hat{\pi}\}}
   \left| \mathbb{E}_\pi \left[ 
   \sum_{h=1}^H \left( Q(s_h,a_h) - (\mathcal{T}_\beta Q)(s_h,a_h) \right) 
   \right] \right|
\end{equation}
The soft performance difference lemma establishes that minimizing the TBRM objective directly compresses the suboptimality gap, mathematically guaranteeing policy convergence toward the global optimum from arbitrary off-policy data. In this work, we build directly upon the TBRM framework introduced by \citet{yuan2026trajectory}; we refer the reader to their work for expanded theoretical derivations. 

\subsection{Reinforce Leave-One-Out (RLOO) Grouping Strategy} 
While policy-gradient methods like GRPO utilize group-relative reward normalization to mitigate gradient variance, value-based trajectory objectives require variance control centered around their residual errors. We introduce a modified Reinforce Leave-One-Out (RLOO) grouping strategy that leverages collective prompt statistics. Crucially, the adjustment is performed using Bellman residuals rather than raw rewards.

For each unique prompt group $g \in \{1, \dots, G\}$, we sample $b \in \{1, \dots, n_g\}$ parallel trajectory rollouts. We define the raw trajectory Bellman error $E_{b,g}$ for response $b$ within group $g$ as:
\begin{equation}
E_{b,g} = V_\theta(s_1^{(b,g)}) - \frac{r(\tau^{(b,g)})}{\beta} + \sum_{h=1}^H \log \frac{\pi_\theta(a_h^{(b,g)}|s_h^{(b,g)})}{\pi_{\text{ref}}(a_h^{(b,g)}|s_h^{(b,g)})}
\end{equation}
To minimize variance without introducing out-of-sample bias, each rollout's residual is adjusted against a counterfactual baseline computed from the remaining samples within its prompt group:
\begin{equation}
\tilde{E}_{b,g} = E_{b,g} - \frac{1}{n_g-1}\sum_{i \neq b} E_{i,g} 
\end{equation}
This baseline measures the residual variance relative to the group mean, thereby facilitating more stable optimization of the neural network. To prevent optimization drifting while maintaining low variance, we optimize a combinational loss function that penalizes both group-relative and absolute residuals:
\begin{equation}
\mathcal{L}_{\text{total}} = \alpha \cdot \mathcal{L}_{\text{RLOO}} + (1-\alpha) \cdot \mathcal{L}_{\text{absolute}}
\end{equation}
where $\mathcal{L}_{\text{RLOO}}$ and $\mathcal{L}_{\text{absolute}}$ are defined over a prompt batch size $B$ and group size $G$ as:
\begin{align}
\mathcal{L}_{\text{RLOO}} &= \frac{1}{B \cdot G} \sum_{b=1}^B \sum_{g=1}^G (\tilde{E}_{b,g})^2 \\
\mathcal{L}_{\text{absolute}} &= \frac{1}{B \cdot G} \sum_{b=1}^B \sum_{g=1}^G (E_{b,g})^2
\end{align}
The weighting hyperparameter $\alpha$ is tuned within the stable regime of $[0.8, 0.99]$. By standardizing optimization over group-relative residuals rather than rewards, this TBRM-RLOO configuration structurally out-performs GRPO-enhanced implementations of TBRM.

\subsection{Hindsight Goal Relabeling} 
In complex multi-step arithmetic tasks, random generation during early post-training phases leads to a high frequency of failed rollouts. These failed trajectories yield persistent zero rewards, leading to uninformative gradient updates and slow optimization. To alleviate this sample inefficiency, we integrate Hindsight Goal Relabeling (HGR).

HGR systematically converts failed trajectories into valuable learning opportunities via the following sequence:
\begin{enumerate}
     \renewcommand{\labelenumi}{\arabic{enumi})}
    \item \textbf{Extraction:} Trajectories that fail to satisfy the original terminal condition or target goal are filtered from the active generation batch.
    \item \textbf{Relabeling:} The actual outcome produced by the model's multi-step derivation is evaluated. The target goal within the initial prompt context is then ex-post relabeled to match this actual achieved outcome, effectively converting a failed rollout into a successful demonstration for a modified task instance.
    \item \textbf{Augmentation:} The newly relabeled valid samples are combined with the original successful trajectories, inflating the coverage of the active replay buffer.
\end{enumerate}
The augmented dataset is passed directly into the group TBRM-RLOO optimization pipeline, ensuring that the model extracts meaningful token-level supervision signals even from sub-optimal rollouts, drastically increasing sample efficiency.

\subsection{$\beta$ Hyperparameter and Algorithmic Procedure}
The temperature hyperparameter $\beta$ controls the trade-off between maximizing the raw reward signal and remaining regularized against the reference policy $\pi_{\text{ref}}$:
\begin{itemize}
    \item \textbf{High $\beta$ Regularization:} Excessively high values over-dampen the reward scale and minimize the impact of optimization updates, leading to slow training velocity, objective instability, and failure to reach convergence.
    \item \textbf{Low $\beta$ Regularization:} Overly aggressive low values amplify the reward scale and optimization updates, causing the policy to overfit rapidly to sparse signals, collapsing the distribution variance and degrading generation quality.
\end{itemize}
For intensive token-level arithmetic reasoning tasks such as the Countdown math puzzle, configuring $\beta$ within the range of $[0.01, 0.02]$ provides the optimal boundary to maximize learning velocity while preventing model collapse. 

The complete training sequence is formalized in Algorithm \ref{alg:tbrm_rloo}.

\begin{algorithm}[H]
\caption{Bellman-Consistent Group TBRM-RLOO with Hindsight Relabeling}
\label{alg:tbrm_rloo}
\begin{algorithmic}[1]
\Require Task dataset $\mathcal{D}$, reference model $\pi_{\text{ref}}$ with weights $\theta_{\text{ref}}$, temperature $\beta$, combination coefficient $\alpha$, group rollout size $G$, iterations $T$, group rollout size $G$.
\State Initialize model weights $\theta \leftarrow \theta_{\text{ref}}$
\State Initialize global step: $\text{global\_step} \leftarrow 0$
\State Initialize storage: $\mathcal{D}_{\text{batch}} \leftarrow \emptyset$ and $\mathcal{D}_{\text{relabel}} \leftarrow \emptyset$

\While{$\text{global\_step} < T$}
    
    \Statex \Comment{\textbf{Phase 1: Rollout Generation \& Hindsight Relabeling}}
    \If{$|\mathcal{D}_{\text{batch}}| = 0$ \textbf{and} $|\mathcal{D}_{\text{relabel}}| = 0$} 
        \If{$\mathcal{D}$ iterator is exhausted}
            \State Reset dataset iterator $\mathcal{D}$
        \EndIf
        \State $\mathcal{D}_{\text{batch}} \leftarrow \text{next batch from } \mathcal{D}$
     
        \For{each prompt $q \in \mathcal{D}_{\text{batch}}$}
            \State Generate $G$ parallel rollouts $\tau^{(1)}, \dots, \tau^{(G)} \sim \pi_\theta(\cdot|q)$
            \For{each rollout $\tau^{(i)}$}
                \If{$\tau^{(i)}$ fails target goal $g_{\text{orig}}$}
                    \State Extract actual outcome $g_{\text{actual}}$ from terminal state
                    \State Relabel prompt $q \rightarrow q_{\text{new}}$ using $g_{\text{actual}}$
                    \State Collect to relabel pool: $\mathcal{D}_{\text{relabel}} \leftarrow \mathcal{D}_{\text{relabel}} \cup \{\tau^{(i)}_{\text{relabeled}}\}$
                \EndIf
            \EndFor
        \EndFor

    \Statex \Comment{\textbf{Phase 2: Train on Standard Batch Data}}
    \ElsIf{$|\mathcal{D}_{\text{batch}}| > 0$}
        \State Rollout $G$ generation for each prompt in $\mathcal{D}_{\text{batch}}$ 
        \State Compute Bellman residuals $E_{b,g}$ and RLOO-adjusted residuals $\tilde{E}_{b,g}$ for $\mathcal{D}_{\text{batch}}$
        \State Evaluate combinational loss $\mathcal{L}_{\text{total}} = \alpha \mathcal{L}_{\text{RLOO}} + (1-\alpha) \mathcal{L}_{\text{absolute}}$
        \State Update parameters: $\theta \leftarrow \theta - \eta \nabla_\theta \mathcal{L}_{\text{total}}$
        \State Clear batch data: $\mathcal{D}_{\text{batch}} \leftarrow \emptyset$
        \State Increment global step: $\text{global\_step} \leftarrow \text{global\_step} + 1$

    \Statex \Comment{\textbf{Phase 3: Retrain using Hindsight Relabeled Data (Multi-step Pop)}}
    \ElsIf{$|\mathcal{D}_{\text{batch}}| = 0$ \textbf{and} $\mathcal{D}_{\text{relabel}} \neq \emptyset$}
        \State Pop a mini-batch $\mathcal{B}_{\text{relabel}} \subset \mathcal{D}_{\text{relabel}}$
        \State Rollout $G$ generation for each prompt in $\mathcal{B}_{\text{relabel}}$ 
        \State Compute Bellman residuals $E_{b,g}$ and RLOO-adjusted residuals $\tilde{E}_{b,g}$ for $\mathcal{B}_{\text{relabel}}$
        \State Evaluate combinational loss $\mathcal{L}_{\text{relabel}} = \alpha \mathcal{L}_{\text{RLOO}} + (1-\alpha) \mathcal{L}_{\text{absolute}}$
        \State Retrain/Update parameters: $\theta \leftarrow \theta - \eta \nabla_\theta \mathcal{L}_{\text{relabel}}$
        \State Remove trained samples from pool: $\mathcal{D}_{\text{relabel}} \leftarrow \mathcal{D}_{\text{relabel}} \setminus \mathcal{B}_{\text{relabel}}$
    \EndIf
\EndWhile
\end{algorithmic}
\end{algorithm}

\section{Experimental Setup}
All fine-tuning experiments were executed on the Modal cloud platform (\url{https://modal.com}) utilizing NVIDIA H100 GPU instances. The software framework was implemented within a Conda environment using native PyTorch code, leveraging the Hugging Face \texttt{transformers} and \texttt{datasets} libraries. Depending on the optimization paradigm, two distinct model initializations were utilized. For SFT, IPO and DPO training configurations, the foundational base model was \texttt{Qwen/Qwen2.5-0.5B}. Conversely, for the online reinforcement learning methods—specifically RLOO, FlowRL, and our proposed framework—training commenced from the recommended fine-tuned checkpoint \texttt{asingh15/qwen-sft-countdown-defaultproj} available on Hugging Face.

The default Countdown pipeline integrates three distinct public datasets from Hugging Face, each serving a tailored role within our multi-phase training workflow. To initialize the model before alignment, a warm-start SFT phase is performed using the strategy dataset (\url{https://huggingface.co/datasets/Asap7772/cog_behav_all_strategies}). For offline, preference-based optimization paradigms that contrast paired trajectories—such as DPO—we employ the pairwise preference dataset (\url{https://huggingface.co/datasets/asingh15/countdown_tasks_3to4-dpo}). Finally, online verifier-based reinforcement learning methods (such as RLOO, flowRL and TBRM) and overall tracking utilize the prompt and evaluation dataset (\url{https://huggingface.co/datasets/asingh15/countdown_tasks_3to4}), maintaining its default training and testing partitions in the default project. Within this online RL regime, prompts are extracted from the dataset to generate parallel rollouts, which are subsequently evaluated and scored by a deterministic, rule-based countdown verifier.

During the optimization phase, models are generally trained for a maximum horizon of 200 steps. Model validation is performed continuously, and the optimal parameters are checkpointed and saved based on the highest achieved rollout mean accuracy. To mitigate overfitting and minimize computational waste, an early stopping criterion is enforced, which automatically terminates the training process if the rollout mean accuracy fails to improve for 20 consecutive evaluation iterations.

\section{Results}
\subsection{Quantitative Evaluation}

\subsubsection{\texorpdfstring{$\beta$}{Beta} Hyperparameter Exploration}
\begin{figure}[htbp]
  \centering
  \includegraphics[width=0.8\linewidth]{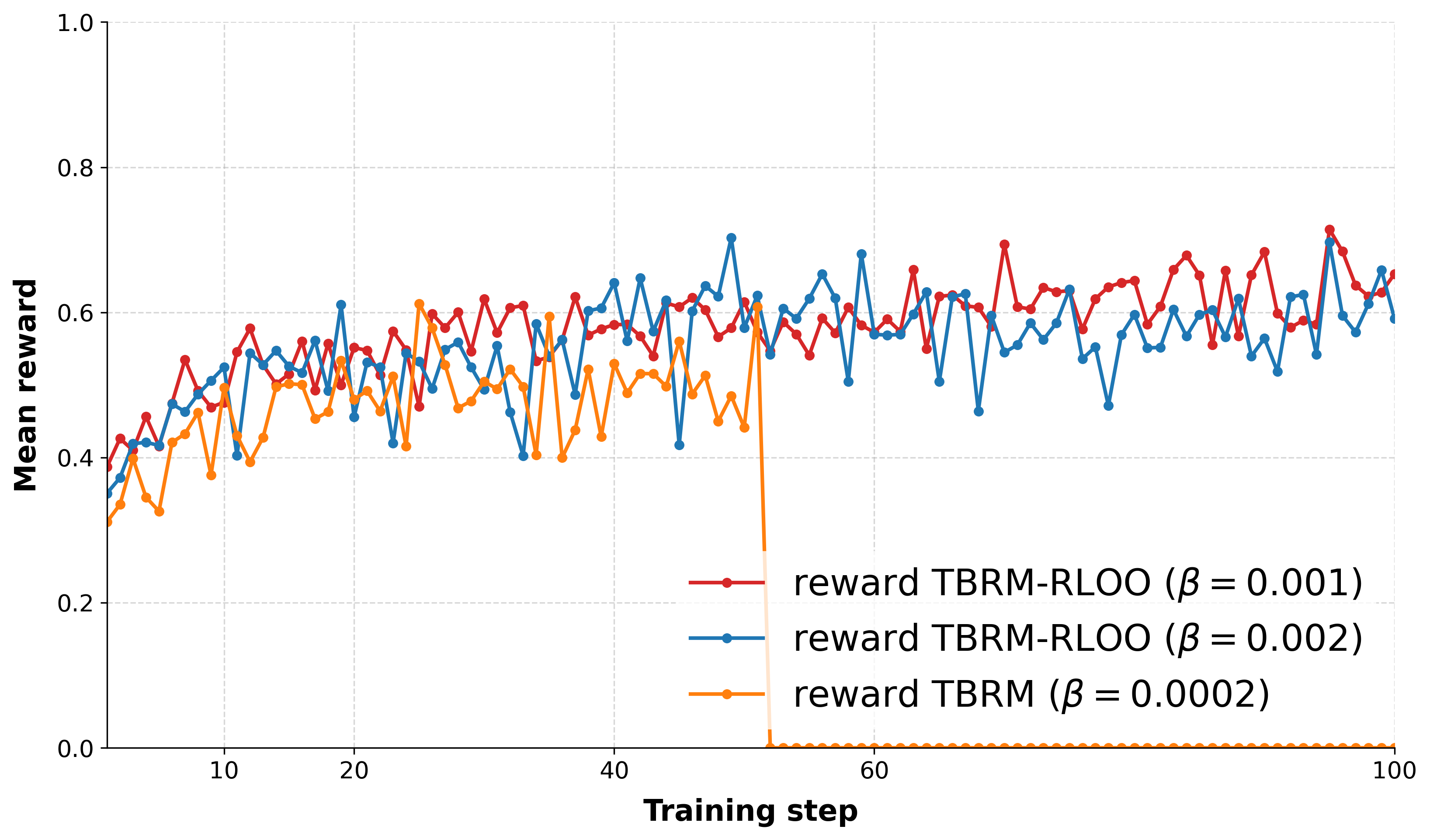}
  \caption{Empirical evaluation of the impact of the temperature hyperparameter $\beta$ on optimization stability and reward scaling.}
  \label{fig:res_beta}
\end{figure}

The temperature hyperparameter, $\beta$, serves as a critical regularizer governing the trade-off between reward maximization and policy divergence. As illustrated in Figure~\ref{fig:res_beta}, the magnitude of $\beta$ dictates optimization stability. An excessively high $\beta$ over-penalizes deviations from the reference policy, effectively dampening the relative reward scale and stalling optimization progress; under extreme settings, this over-regularization induces training instability and prevents convergence. 

Conversely, a small $\beta$ value amplifies the policy gradient updates by scaling up the effective reward variance, yet values that are too aggressive precipitating catastrophic policy collapse. For our specific countdown task, empirical tuning demonstrates that $\beta = 0.001$ consistently yields the optimal balance between exploration and stable policy updates.

\subsubsection{Learning Curve Analysis}
\begin{figure}[htbp]
  \centering
  \includegraphics[width=0.8\linewidth]{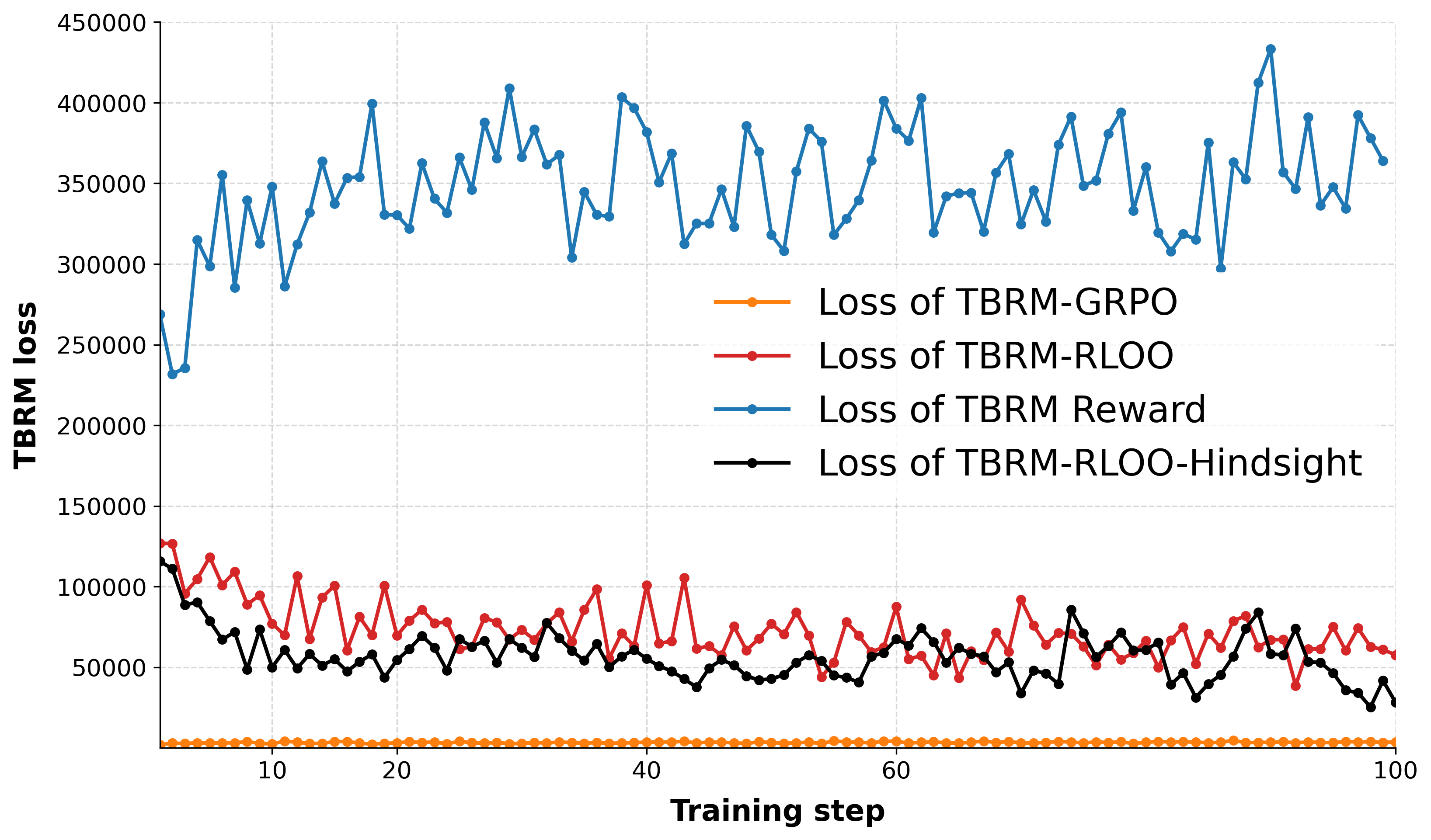}
  \caption{Comparison of Bellman residual loss learning curves across the four evaluated TBRM variants.}
  \label{fig:res_loss}
\end{figure}

We evaluated the training dynamics of four distinct variants: vanilla TBRM, TBRM-GRPO, TBRM-RLOO, and TBRM-RLOO enhanced with hindsight relabeling. The Bellman residual loss trajectories presented in Figure~\ref{fig:res_loss} reveal that vanilla TBRM exhibits severe high-variance oscillations and fails to establish a clear downward trend, underscoring the instability of unconstrained Bellman residual minimization. 

In contrast, incorporating REINFORCE Leave-One-Out (RLOO) significantly stabilizes training, mitigating variance and producing a steady, monotonic decline in the residual loss. Furthermore, integrating hindsight relabeling (TBRM-RLOO-Hindsight) yields the most favorable optimization profile, characterized by minimized baseline loss and accelerated variance reduction compared to standard RLOO execution. 

Notably, GRPO-TBRM induces an artificially suppressed loss magnitude. The group-relative advantage normalization inherent to GRPO over-flattens the gradient variance across samples, which ultimately stymies robust representation learning and leads to premature stagnation.
 
\begin{figure}[htbp]
  \centering
  \includegraphics[width=0.8\linewidth]{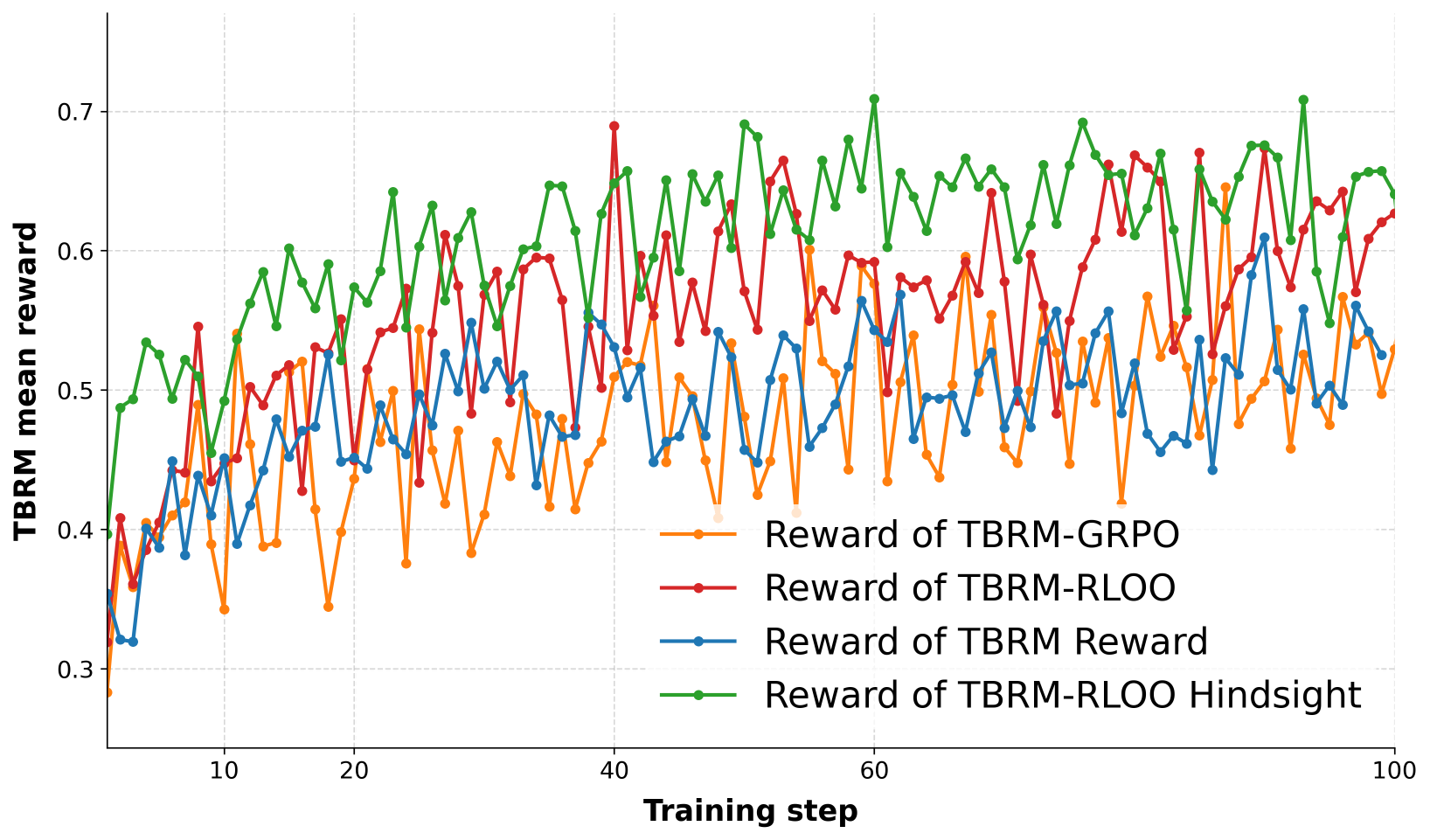}
  \caption{Roll-out reward score trajectories during training across different TBRM configurations.}
  \label{fig:res_reward}
\end{figure}

These optimization behaviors are closely mirrored by the roll-out reward scores during training (Figure~\ref{fig:res_reward}). TBRM-RLOO with hindsight relabeling exhibits the most rapid and superior reward trajectory among all configurations. Additionally, the baseline TBRM-RLOO maintains a consistently higher reward profile than both the vanilla and GRPO-regularized variants, validating the efficacy of the variance-reduced policy gradient formulation.  

\subsubsection{Comprehensive Performance Comparison}
\begin{figure}[htbp]
  \centering
  \includegraphics[width=0.9\linewidth]{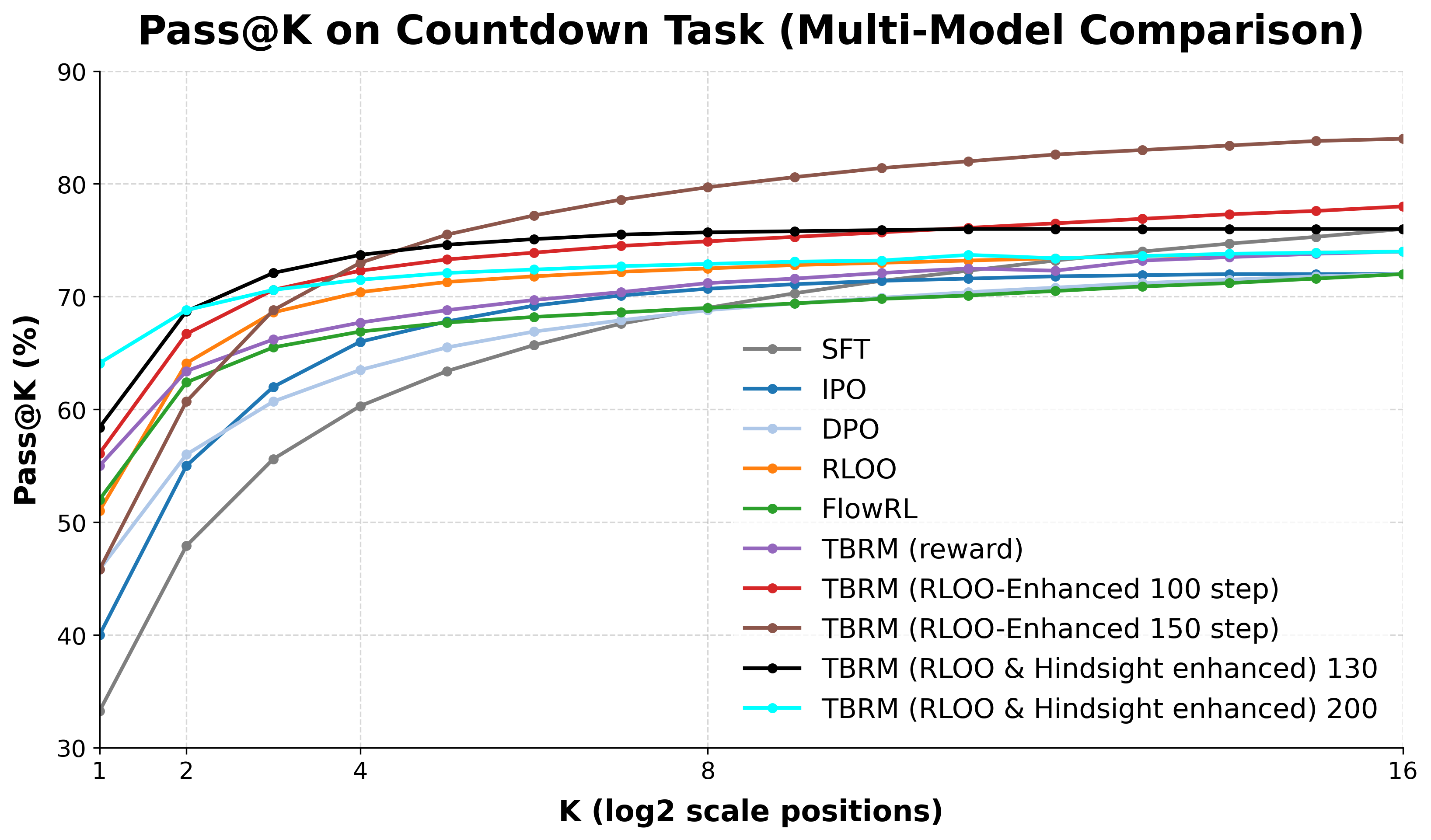}
  \caption{Comprehensive Pass@k performance comparison across baseline architectures and extension models.}
  \label{fig:passatk}
\end{figure}

Table~\ref{tab:performance} and Figure~\ref{fig:passatk} summarize the global performance metrics across baselines and proposed methods evaluated under identical compute budgets.

\begin{table}[htbp]
  \caption{Performance Comparison Across Baseline and Proposed Methods}
  \label{tab:performance}
  \centering
  \begin{adjustbox}{max width=\textwidth}
  \renewcommand{\theadfont}{\normalfont}
  \begin{tabular}{lcccccc}
    \toprule
    Method & \thead[l]{Rollout\\Accuracy$^*$} & Pass@1 & Pass@3 & Pass@6 & Pass@10 & Pass@16 \\
    \midrule
    SFT Baseline & - & 0.33 & 0.55 & 0.65 & 0.71 & 0.76 \\
    IPO          & - & 0.40 & 0.62 & 0.69 & 0.71 & 0.72 \\
    DPO          & - & 0.45 & 0.61 & 0.67 & 0.70 & 0.72 \\
    RLOO         & 0.66 & 0.51 & 0.69 & 0.72 & 0.73 & 0.74 \\
    \midrule
    \multicolumn{7}{l}{\textit{Extension Models}} \\
    FlowRL       & 0.70 & 0.52 & 0.66 & 0.68 & 0.70 & 0.72 \\
    TBRM (reward)& 0.61   & 0.55 & 0.66 & 0.70 & 0.72 & 0.74 \\
    TBRM-RLOO (100 step)         & 0.70 & 0.56 & 0.71 & 0.74 & 0.76 & 0.78 \\
    TBRM-RLOO (150 step)         & 0.73 & 0.46 & 0.69 & 0.77 & 0.81 & 0.84 \\
    TBRM-RLOO-Hindsight (130 step)  & 0.71 & 0.58 & 0.72 & 0.75 & 0.76 & 0.76 \\
    TBRM-RLOO-Hindsight (200 step)  & 0.71 & 0.64 & 0.71 & 0.72 & 0.73 & 0.74 \\
    \bottomrule
    \multicolumn{7}{l}{\footnotesize $^*$ Reported highest accuracy reached during training.} \\
  \end{tabular}
  \end{adjustbox}
  \vspace{-3pt}
\end{table}

\paragraph{Significant Gains over Vanilla TBRM:} TBRM-RLOO (150 steps) markedly improves rollout accuracy from 0.61 to 0.73 (about 20\% relative improvement) and Pass@16 from 0.74 to 0.84 (about 14\% relative improvement) compared to vanilla TBRM, proving that structured variance reduction translates directly into better test-time generalization.

\paragraph{Superior Peak Generation Accuracy:} TBRM-RLOO at 150 steps achieves the top overall multi-sample performance with a Pass@16 score of 84.0\%. This outpaces the standalone RLOO baseline by 10.0\% (absolute) and the supervised Fine-Tuning (SFT) baseline by 8.0\% (absolute). 

\paragraph{Single-Sample Compute Efficiency:} Within low-compute regimes (Pass@1), TBRM-RLOO at 100 steps yields a strong performance of 56.1\%, outperforming standard baseline models. This single-generation efficiency is only surpassed by configurations integrating hindsight relabeling; notably, TBRM-RLOO-Hindsight at 200 steps achieves a peak Pass@1 score of 64.0\%, highlighting that hindsight trajectories significantly accelerate early-stage sample efficiency.

\paragraph{Surpassing the Alignment Ceiling:} Traditional alignment frameworks (such as DPO and IPO) plateau early, capping performance near 72.0\% on Pass@16. TBRM breaks through this apparent bottleneck, leveraging test-time trajectory modeling to continuously scale accuracy with extended training steps.

Crucially, the data indicates that hindsight relabeling acts as an effective catalyst during the initial phases of training, accelerating early reward acquisition and driving substantial gains in intermediate representation learning capacity.

\subsection{Qualitative Analysis}
To investigate the structural shifts in model behavior and the underlying reasoning patterns induced by our framework, we conducted a qualitative analysis on randomly sampled generation test cases from the evaluation set. 

\begin{figure}[t!]
  \centering
  \includegraphics[width=1\linewidth]{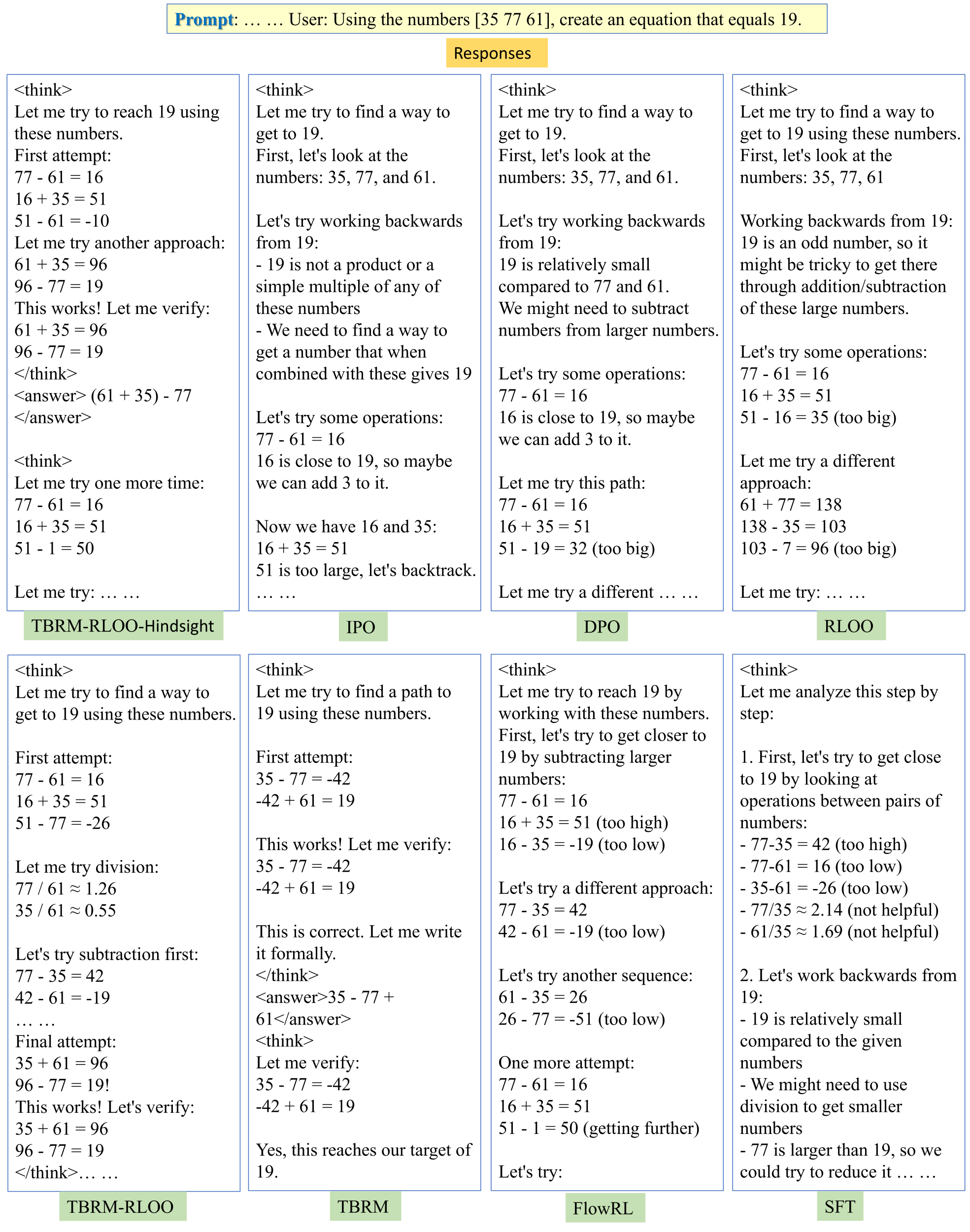}
  \caption{Case study comparison of reasoning path generations for a representative countdown task.}
  \label{fig:res_response_case}
\end{figure}

As illustrated by the representative countdown task instance in Figure~\ref{fig:res_response_case}, TBRM demonstrates a clear capacity to construct more concise, streamlined mathematical reasoning paths to arrive at the correct target value compared to standard baseline alternatives. Incorporating the hybrid TBRM-RLOO framework with hindsight relabeling also optimizes this trajectory, significantly mitigating logical redundancy and enabling the agent to satisfy numeric constraints using fewer intermediate operational steps. 

In contrast, offline preference-based alignment methods like IPO and DPO generate highly redundant inference sequences. While online RLOO reduces the total number of predicted tokens, its outputs still retain substantial structural redundancy, though they remain markedly superior to the unaligned SFT baseline. Meanwhile, FlowRL produces shorter individual responses but frequently requires multiple circuitous internal attempts within its generation token sequence to ultimately converge on the correct answer.

These qualitative observations demonstrate that TBRM successfully propagates terminal reward signals backward to intermediate decision nodes. Furthermore, the integration of hindsight relabeling strengthens this credit assignment mechanism, optimizing step-by-step intermediate logic and effectively discouraging mathematically circuitous generation paths.

\section{Discussion}
This project introduces a Bellman-consistent trajectory reinforcement learning framework designed to enhance intermediate learning signals for large language model mathematical inference, specifically evaluated on the countdown task. By leveraging TBRM, our approach successfully backpropagates terminal reward signals to individual intermediate inference steps, thereby systematically optimizing the reasoning chain. Distinct from traditional token-level policy optimization methods, the theoretical foundation of TBRM relies on the soft performance difference lemma. This ensures that the framework converges toward a near-optimal solution by minimizing Bellman-consistent residuals, effectively transforming sparse, terminal rewards into dense, token-level credit assignments. Empirically, we observed that the temperature hyperparameter $\beta$ plays a critical role in stabilizing policy optimization. Our behavioral analysis confirms that TBRM successfully prunes redundant reasoning steps, encouraging the model to discover concise and direct paths to the target value.

To address the inherent instability of policy gradient methods in language generation, we introduced two key enhancements:
\begin{itemize}
    \item \textbf{Robust Variance Reduction via RLOO:} Incorporating the REINFORCE Leave-One-Out (RLOO) estimator to adjust Bellman residuals drastically reduces gradient variance. In mathematical reasoning tasks, context and prompt difficulty vary widely, typically causing severe reward fluctuations. RLOO effectively normalizes these baseline discrepancies across trajectories, establishing a stable, steadily declining loss curve.
    \item \textbf{Sample Efficiency via Hindsight Experience Replay:} Hindsight goal relabeling acts as a powerful data augmentation mechanism to reinforce suboptimal or failed trajectories. Our experimental results demonstrate a clear performance hierarchy: while vanilla TBRM suffers from high variance, and TBRM-RLOO stabilizes training, the integration of hindsight relabeling yields the fastest convergence, the lowest loss variance, and the higher overall reward curves.
\end{itemize}

\subsection{Limitations and Future Work}
Despite these promising results, several limitations remain to be addressed in future research. First, we observe a distinct trade-off in policy behavior over extended optimization horizons. As shown in Table~\ref{tab:performance}, increasing the training steps of the RLOO-TBRM method from 100 to 150 steps results in a degradation in greedy decoding performance (Pass@1 dropping from 0.56 to 0.46) alongside a substantial improvement in exploratory coverage (Pass@16 rising from 0.78 to 0.84). This pattern indicates that while extended optimization enhances the diversity and maximum capacity of the sampled reasoning paths, it simultaneously degrades top-1 deterministic precision—potentially due to high policy variance or distribution drift during late-stage reinforcement learning. Investigating the underlying mechanics of this trade-off to preserve greedy precision while retaining wide coverage remains an immediate priority.

Furthermore, we observed a susceptibility to late-stage mode collapse when training the trajectory-based TBRM-RLOO-Hindsight configuration over extended horizons beyond 200 steps. This phenomenon is likely driven by the policy overfitting to synthetic, relabeled target goals. To mitigate this structural collapse, future iterations will explore dynamic curriculum scheduling mechanisms. Such schedules will gradually decay or balance the ratio between original task distributions and hindsight-relabeled trajectories as training progresses, preventing the policy from destabilizing on out-of-distribution synthetic objectives.

Hyperparameter sensitivity and architectural constraints also present opportunities for expansion. Currently, the local policy temperature parameter $\beta$ is held static; however, given its heavy influence on optimization stability, transitioning $\beta$ into an adaptive, dynamically learnable parameter warrants systematic exploration. Mechanistically, embedding this unified TBRM framework within a Mixture of Experts (MoE) architecture could allow the model to dynamically route and fuse heterogeneous loss objectives (e.g., balancing RLOO and trajectory Bellman residuals) conditioned on the complexity of the prompt. Finally, while the current empirical evaluation demonstrates strong validation on the Countdown math puzzle, expanding the benchmark scope to more complex, open-ended multi-step reasoning environments such as GSM8K and MATH is essential to rigorously evaluate the cross-domain generalization and scalability of our framework.

\section{Conclusion}
This paper presented a Bellman-consistent trajectory reinforcement learning framework designed to address the pervasive challenge of sparse rewards in LLM-based mathematical reasoning. By leveraging the existing methodology of TBRM, we successfully map sparse terminal rewards into dense credit assignments along the inference trajectory. Grounded in the soft performance difference lemma, our framework builds upon the theoretical guarantees of TBRM—specifically, that minimizing the trajectory-level Bellman residual drives the policy toward a near-optimal solution. To overcome the optimization instabilities and sample inefficiencies that arise when applying this trajectory method to sequential language generation, we integrated the REINFORCE Leave-One-Out (RLOO) estimator for robust variance reduction alongside hindsight goal relabeling for data augmentation.

Our empirical evaluations demonstrate that the unified TBRM-RLOO-Hindsight framework significantly outperforms baseline token-level policy optimization methods, delivering accelerated convergence, gradient variance reduction, and a marked reduction in reasoning redundancy. While future work is required to alleviate late-stage mode collapse and extend validation to broader mathematical benchmarks such as GSM8K and MATH, this work validates the practical efficacy of trajectory-level credit assignment for language models. Future extensions exploring dynamically learnable hyperparameters and Mixture-of-Experts (MoE) architectures will further unlock the generalizability of Bellman-consistent policy optimization for complex, multi-step cognitive tasks.

\section{Team Contributions}
As the sole investigator, \textbf{Lianfa Li} is responsible for the execution and management of all aspects of this project, structured into four core operational phases:

\begin{itemize}
    \item \textbf{Environment Setup \& Data Preparation:} Configure the Countdown mathematical inference environment, curate training and evaluation datasets, and establish baseline evaluation setups for token-level policy optimization methods and standalone TBRM variants.
    
    \item \textbf{Algorithmic Framework Design:} Develop the unified Bellman-consistent trajectory reinforcement learning framework, combining the TBRM loss formulation with the REINFORCE Leave-One-Out (RLOO) variance reduction estimator and a hindsight experience replay mechanism via goal relabeling.
    
    \item \textbf{Implementation \& Empirical Evaluation:} Implement the complete algorithmic pipeline utilizing Qwen as the base large language model (LLM), execute systematic ablation studies to isolate the contributions of TBRM, RLOO, and hindsight data augmentation, and track training metrics including Pass@$k$, loss variance, and cumulative reward curves.
    
    \item \textbf{Analysis \& Comprehensive Reporting:} Profile policy training convergence and late-stage mode collapse dynamics, perform behavioral analyses regarding reasoning step redundancy pruning, evaluate the sensitivity of the temperature hyperparameter $\beta$, and compose the final research manuscript.
\end{itemize} 

\paragraph{Changes from Proposal} 
While the core objective of transforming sparse terminal rewards into dense token-level feedback via TBRM remains fully aligned with the original proposal, the specific variance reduction and sample efficiency mechanisms were adapted during empirical iteration. 

First, rather than utilizing Group Relative Policy Optimization (GRPO) for trajectory-level advantage estimation as initially planned, we integrated the REINFORCE Leave-One-Out (RLOO) estimator applied directly to the Bellman residuals. This variation provides a more structurally sound and targeted variance reduction mechanism by effectively normalizing baseline discrepancies caused by varying prompt difficulties. 

Second, we replaced the proposed prioritized historical trajectory replay mechanism and its associated off-policy importance-weight corrections with hindsight goal relabeling (Hindsight Experience Replay). For the sequential structures inherent to the Countdown arithmetic task, hindsight relabeling proved to be a far more appropriate data augmentation strategy, allowing the framework to directly reinforce failed or suboptimal trajectories by matching them to synthetic goals. These architectural modifications evolved naturally from the iterative nature of the research and development process, ultimately yielding better training stability and faster convergence.

\bibliographystyle{ACM-Reference-Format}
\bibliography{reference}

@misc{rlhf,
      title={Training language models to follow instructions with human feedback}, 
      author={Long Ouyang and Jeff Wu and Xu Jiang and Diogo Almeida and Carroll L. Wainwright and Pamela Mishkin and Chong Zhang and Sandhini Agarwal and Katarina Slama and Alex Ray and John Schulman and Jacob Hilton and Fraser Kelton and Luke Miller and Maddie Simens and Amanda Askell and Peter Welinder and Paul Christiano and Jan Leike and Ryan Lowe},
      year={2022},
      eprint={2203.02155},
      archivePrefix={arXiv},
      primaryClass={cs.CL}
}

@article{zhu2025flowrl,
  title={Flowrl: Matching reward distributions for llm reasoning},
  author={Zhu, Xuekai and Cheng, Daixuan and Zhang, Dinghuai and Li, Hengli and Zhang, Kaiyan and Jiang, Che and Sun, Youbang and Hua, Ermo and Zuo, Yuxin and Lv, Xingtai and others},
  journal={arXiv preprint arXiv:2509.15207},
  year={2025}
}

@inproceedings{lightman2024let,
  title={Let's verify step by step},
  author={Lightman, Hunter and Kosaraju, Vineet and Burda, Yuri and Edwards, Harrison and Baker, Bowen and Lee, Teddy and Leike, Jan and Schulman, John and Sutskever, Ilya and Cobbe, Karl},
  booktitle={International Conference on Learning Representations},
  volume={2024},
  pages={39578--39601},
  year={2024}
}

@article{andrychowicz2017hindsight,
  title={Hindsight experience replay},
  author={Andrychowicz, Marcin and Wolski, Filip and Ray, Alex and Schneider, Jonas and Fong, Rachel and Welinder, Peter and McGrew, Bob and Tobin, Josh and Pieter Abbeel, OpenAI and Zaremba, Wojciech},
  journal={Advances in neural information processing systems},
  volume={30},
  year={2017}
}

@inproceedings{ahmadian2024back,
  title={Back to basics: Revisiting REINFORCE-style optimization for learning from human feedback in LLMs},
  author={Ahmadian, Arash and Cremer, Chris and Gall{\'e}, Matthias and Fadaee, Marzieh and Kreutzer, Julia and Pietquin, Olivier and {\"U}st{\"u}n, Ahmet and Hooker, Sara},
  booktitle={Proceedings of the 62nd Annual Meeting of the Association for Computational Linguistics (Volume 1: Long Papers)},
  pages={12248--12267},
  year={2024}
}

@article{williams1992simple,
  title={Simple statistical gradient-following algorithms for connectionist reinforcement learning},
  author={Williams, Ronald J},
  journal={Machine learning},
  volume={8},
  number={3},
  pages={229--256},
  year={1992},
  publisher={Springer}
}

@article{rafailov2023direct,
  title={Direct Preference Optimization: Your Language Model is Secretly a Reward Model},
  author={Rafailov, Rafael and Sharma, Archit and Mitchell, Eric and others},
  journal={NeurIPS},
  year={2023}
}

@article{baird1995residual,
  title={Residual Algorithms: Reinforcement Learning with Function Approximation},
  author={Baird, Leemon},
  journal={ICML},
  year={1995}
}

@article{sutton1998reinforcement,
  title={Reinforcement Learning: An Introduction},
  author={Sutton, Richard S. and Barto, Andrew G.},
  journal={MIT Press},
  year={1998}
}

@article{antos2008learning,
  title={Learning near-optimal policies with Bellman-residual minimization based fitted policy iteration},
  author={Antos, Andras and Szepesvari, Csaba and Munos, Remi},
  journal={Machine Learning},
  year={2008}
}

@inproceedings{dann2014policy,
  title={Policy Evaluation with Temporal Differences: A Survey and Comparison},
  author={Dann, Christoph and Neumann, Gerhard and Peters, Jan},
  booktitle={JMLR},
  year={2014}
}

@article{shao2024deepseekmath,
  title={Deepseekmath: Pushing the limits of mathematical reasoning in open language models},
  author={Shao, Zhihong and Wang, Peiyi and Zhu, Qihao and Xu, Runxin and Song, Junxiao and Bi, Xiao and Zhang, Haowei and Zhang, Mingchuan and Li, YK and Wu, Yang and others},
  journal={arXiv preprint arXiv:2402.03300},
  year={2024}
}

@article{schulman2017proximal,
  title={Proximal policy optimization algorithms},
  author={Schulman, John and Wolski, Filip and Dhariwal, Prafulla and Radford, Alec and Klimov, Oleg},
  journal={arXiv preprint arXiv:1707.06347},
  year={2017}
}

@article{ouyang2022training,
  title={Training language models to follow instructions with human feedback},
  author={Ouyang, Long and Wu, Jeffrey and Jiang, Xu and Almeida, Diogo and Wainwright, Carroll and Mishkin, Pamela and Zhang, Chong and Agarwal, Sandhini and Slama, Katarina and Ray, Alex and others},
  journal={Advances in neural information processing systems},
  volume={35},
  pages={27730--27744},
  year={2022}
}

@article{guo2025deepseek,
  title={Deepseek-r1: Incentivizing reasoning capability in llms via reinforcement learning},
  author={Guo, Daya and Yang, Dejian and Zhang, Haowei and Song, Junxiao and Wang, Peiyi and Zhu, Qihao and Xu, Runxin and Zhang, Ruoyu and Ma, Shirong and Bi, Xiao and others},
  journal={arXiv preprint arXiv:2501.12948},
  year={2025}
}

@article{li2025system,
  title={From system 1 to system 2: A survey of reasoning large language models},
  author={Li, Zhong-Zhi and Zhang, Duzhen and Zhang, Ming-Liang and Zhang, Jiaxin and Liu, Zengyan and Yao, Yuxuan and Xu, Haotian and Zheng, Junhao and Wang, Pei-Jie and Chen, Xiuyi and others},
  journal={arXiv preprint arXiv:2502.17419},
  year={2025}
}

@article{herr2025llm,
  title={LLM-first search: Self-guided exploration of the solution space},
  author={Herr, Nathan and Rockt{\"a}schel, Tim and Raileanu, Roberta},
  journal={arXiv preprint arXiv:2506.05213},
  year={2025}
}

@article{katz2025seemingly,
  title={Seemingly Simple Planning Problems are Computationally Challenging: The Countdown Game},
  author={Katz, Michael and Kokel, Harsha and Sreedharan, Sarath},
  journal={arXiv preprint arXiv:2508.02900},
  year={2025}
}

@article{lian2025reinforcement,
  title={Reinforcement Learning is all You Need},
  author={Lian, Yongsheng},
  journal={arXiv preprint arXiv:2503.09512},
  year={2025}
}

@article{yuan2026trajectory,
  title={Trajectory bellman residual minimization: A simple value-based method for llm reasoning},
  author={Yuan, Yurun and Chen, Fan and Jia, Zeyu and Rakhlin, Alexander and Xie, Tengyang},
  journal={Advances in Neural Information Processing Systems},
  volume={38},
  pages={116994--117032},
  year={2026}
}


\end{document}